\documentclass[runningheads]{llncs}
\usepackage{lineno,hyperref}
\modulolinenumbers[5]
\usepackage{graphicx}
\usepackage{tcolorbox}
\usepackage{ulem}
\usepackage{url}
\usepackage{tabularx}
\usepackage{soul}
\usepackage[T1]{fontenc}
\usepackage{commath}
\usepackage{booktabs}
\usepackage{todonotes} 
\begin{document}
\title{Incorporating Joint Embeddings into Goal-Oriented Dialogues with Multi-Task Learning}
\titlerunning{Goal-Oriented Dialogues with Multi-Task Learning}

\author{Firas Kassawat \inst{1}  \and Debanjan Chaudhuri\inst{1, 2} \and Jens Lehmann\inst{1, 2}}

%
\authorrunning{Kassawat, Chaudhuri, Lehmann}
%
\institute{
Smart Data Analytics Group (SDA), University of Bonn, Germany\\{kassawat@cs.uni-bonn.de}\
\{chaudhur, jens.lehmann\}@cs.uni-bonn.de\\
\and
Enterprise Information Systems Department, Fraunhofer IAIS, Bonn, Germany\\
\{debanjan.chaudhuri, jens.lehmann\}@iais.fraunhofer.de
}
%
\maketitle              
\begin{abstract}
Attention-based encoder-decoder neural network models have recently shown promising results in goal-oriented dialogue systems. However, these models struggle to reason over and incorporate state-full knowledge while preserving their end-to-end text generation functionality. Since such models can greatly benefit from user intent and knowledge graph integration, in this paper we propose an RNN-based end-to-end encoder-decoder architecture which is trained with joint embeddings of the knowledge graph and the corpus as input. The model provides an additional integration of user intent along with text generation, trained with multi-task learning paradigm along with an additional regularization technique to penalize generating the wrong entity as output. The model further incorporates a Knowledge Graph entity lookup during inference to guarantee the generated output is state-full based on the local knowledge graph provided. We finally evaluated the model using the BLEU score,  empirical evaluation depicts that our proposed architecture can aid in the betterment of task-oriented dialogue system`s performance.

\end{abstract}
\section{Introduction}
There is a considerable rise in the need for effective task-oriented dialogue agents that can help the user to achieve specific goals and everyday tasks such as weather forecast assistance, schedule arrangement, and location navigation. Neural network based dialogue models proved to be the most promising architectures so far which can effectively use dialogue corpora in order to build such agents. However, these models struggle to reason over and incorporate state-full knowledge while preserving their end-to-end text generation functionality. They often require additional user supervision or additional data annotation to incorporate some dialogue-state information along with additional knowledge.

To help the dialogue agents overcome these problems, they have often been built with several pipelined modules such as language understanding, dialog management, question answering components, and natural language generation~\cite{P18-1136}. However, modeling the dependencies between such modules is complex and may not result in very natural conversations.

The problem becomes much more complicated in multi-domain scenarios, because the text generation in such scenarios hugely depends on the domain the user is talking about, where each domain will have its own vocabulary that differs from other domains. Hence, understanding user intent during the text generation process can be beneficial. Table \ref{tab:intent} shows sample user utterances along with its intent. While generating new tokens, the model can benefit from this because separate intents will follow different vocabulary distributions.

\begin{table*}[ht]
    \caption{User query and respective intents}
     \begin{tabular}{ l | l}
     \toprule
     \textbf{User utterance} & \textbf{Intent}\\
    \hline
     Book a seat for me in any good Chinese restaurant nearby &	Restaurant Booking \\
     Book a flight ticket for me from Berlin to Paris for tomorrow	& Flight Booking \\
     Please show me the direction to the main train station & Navigation \\
     Please cancel my appointment with the dentist for tomorrow & Scheduling \\
     \bottomrule
    \end{tabular}
 \vspace{0.1cm}
    \label{tab:intent}
\end{table*}

In order to tackle the aforementioned problems, we propose a novel neural network architecture trained using a multi-task learning paradigm, which along with generating tokens from the vocabulary also tries to predict the intent of the user utterances. By doing so, the model is able to relate the generated words to the predicted intent and uses this to exclude words unrelated to the current conversation from the next word candidate prediction. Our model utilizes joint text and knowledge graph embeddings as inputs to the model inspired by the works of~\cite{Alsuhaibani2018}.

Additionally, after projecting the entities and text into the same vector space, we propose an additional training criterion to use as a novel regularization technique called entity loss, which further penalizes the model if the predicted entity is different from the original. The loss is the mean of the vector distance between the predicted and correct entities. 



Furthermore, to guarantee a state-full knowledge incorporation we included a knowledge graph (KG) key-value look-up 
to implement a way of knowledge tracking during the utterances of a dialogue.

A KG in the context of this paper is a directed, multi-relational graph that represents entities as nodes, and their relations as edges, and can be used as an abstraction of the real world. KGs consists of triples of the form (h,r,t) \begin{math}\in KG\end{math}, where h and t denote the head and tail entities, respectively, and r denotes their relation.

\noindent
Our main contributions in this paper are as follows:

\begin{itemize}
    \item Utilizing joint text and knowledge graph embeddings into an end-to-end model for improving the performance of task-oriented dialogue systems.
    \item A multi-task learning of dialogue generation and learning user intent during decoding.
    \item A novel entity loss based regularization technique which penalizes the model if the predicted entity is further from the true entity in vector space.
\end{itemize}


\section{Related Work}

Recently, several works have tried incorporating external knowledge, both structured and unstructured, into dialogue systems.
\cite{lowe2015incorporating} \cite{xu2016incorporating} \cite{kosovan-2017-cscubs-dialogues} proposed architectures for incorporating unstructured knowledge into retrieval based dialogue systems targeting the Ubuntu dialogue corpus \cite{DBLP:journals/corr/LowePSP15}. More recently, \cite{chaudhuri2018improving} used domain description based unstructured knowledge using an additional recurrent neural network (GRU) architecture added to the word embeddings for domain keywords to further improve the performance of such models.
There are also many recent works that incorporated end-to-end models with structured knowledge sources, such as knowledge graphs \cite{wen2016network}, \cite{dhingra2016towards}, \cite{williamsend}. \cite{eric-manning:2017:SIGDIAL} used a key-value retrieval network to augment the vocabulary distribution of the keys of the KG with the attentions of their associated values. We were inspired by this model to add the key-value lookup in the inference stage of the model. Although the work they represented was inspiring, it turns out that using attention over a knowledge graph is inefficient for long sequences. Moreover, they didn't utilize user intent during modeling. In~\cite{P18-1136}, the authors used a memory-to-sequence model that uses multi-hop attention over memories to help in learning correlations between memories which results in faster trained model with a stable performance. 

As for using joint learning to support end-to-end dialogue agent the work introduced by~\cite{liuBing2016} showed state-of-the-art results where they used an attention based RNN for the joint learning of intent detection and slot filling. They proposed two methods, an encoder-decoder model with aligned inputs and an attention-Based model. Our intent predictor during decoding is influenced by their architecture. \cite{P17-4013}~implemented a multi-domain statistical dialogue system toolkit in which they created a topic tracker, which is used to load only the domain-specific instances of each dialogue model. In other words, all the elements in the pipeline (semantic decoder, belief tracker and the language generator) would consist of only domain-specific instances.  

Graph embeddings have been used by~\cite{Dettmers2018} to predict missing relationships between entities, later on, \cite{Xiao2017} used knowledge graph embeddings with text descriptions to propose a semantic space projection model that learns from both symbolic triples and their textual description. \cite{alsuhaibani2018jointly}~proposed a model for jointly learning word embeddings using a corpus and a knowledge graph that was used for representing the meaning of words in vector space. They also proposed two methods to dynamically expand a KG. Taking inspirations from these joint embeddings, we incorporate them into our model as inputs for better capturing grounded knowledge.

\section{Model Description}
In this section, we present our model architecture. Firstly, we discuss the attention based RNN encoder-decoder architecture, and then we explain the intent predictor, joint embedding vector space and additional entity loss based regularization.

\subsection{Attention Based Seq-to-seq Model}
Our model is quintessentially based on an RNN-based encoder-decoder architecture taking inspirations from the works proposed by~\cite{vinyals2015neural}, \cite{shang2015neural} and~\cite{luong2014address}. Given the input utterances from the user in a dialogue scenario $D$, the system has to produce an output utterance $o_t$ for each time-step $t$, i.e.~there is a sequence $ \big\{ (i_1, o_1),(i_2, o_2),..., (i_n, o_n) \big\} $ where $n$ denotes the total number of utterances in a dialogue. We reprocess these utterances and present them as:

$ \big\{  (i_1, o_1), ([i_1;o_1;i_2], o_2), ..., ( [i_1; o_1;...;o_{n-1} ;i_n], o_n) \big\} $,    
where we present an input as the sequence of all previous utterances concatenated with the current input. This is done in order to maintain the context in the dialogue.

Let $x_1,...x_m$ denote the words in each input utterance (where $m$ is the length of the utterance). We firstly map these words to their embedding vector representations using $\Phi^{x_t}$ which can be either a randomly initialized or a pretrained embedding. Then these distributed representations are fed into the RNN-encoder (LSTM~\cite{hochreiter1997long} in this case) as follows:
\begin{equation}
h_t = LSTM(\Phi^{{x_t}},h_{t-1})
\end{equation}
\noindent
$h_t=(h_0,....h_m)$ represents the hidden states of the encoder.

These encoder inputs are fed into a decoder which is again a recurrent module. It generates a token for every time-stamp $t$ given by the probability

\begin{equation}
    p(y_t| y_{t-1}, ...,y_1, x_t) = g(y_{t-1}, s_t, c_t)
\end{equation}
\noindent
Where, $g(.)$ is a softmax function, $s_t$ is the hidden state of the decoder at time-step $t$ given by

\begin{equation}
    s_t = LSTM(s_{t-1}, y_{t-1}, c_t)
\end{equation}

Additionally, we use an attention mechanism over the encoder hidden-states as proposed by \cite{bahdanau2014neural}, \cite{graves2015generating}. 
The attention mechanism is calculated as
\begin{equation}
\alpha_{ij}= \frac {exp(e_{ij})}{\sum_{k=1}^m exp(e_{ik})} 
\end{equation}

\begin{equation}
e_{ij} = tanh(W_c[s_{i-1}, h_j])
\end{equation}

The final weighted context from the encoder is given by
\begin{equation}
c_i= \sum_{m} \alpha_{ij} {h_j}
\end{equation}

\noindent
Where, $\alpha_{ij}$ represent the attention weights learned by the model, $c_i$ the context vector, $W_c$ a weight parameter that has to be learned by the model.

\subsection{Predicting Intent}

The intent predictor module is used to predict the intent from the user utterance. It is computed using the encoder LSTM`s hidden representation $h_t$ as computed at every timestamp $t$.
\noindent
The intent output prediction is given by:

\begin{equation}
i_{out}= W_o(tanh(W_i[h_t; c_t])))
\end{equation}

\noindent
Where, $i_{out}$ is the intent score $\in \mathbf{R}^{i_c}$ for the $t^{th}$ time-step, $W_{i}$ and $W_{o}$ are the trainable weight parameters for the intent predictor and $i_c$ is the total number of intent classes. The latter can be the different domains the dialogue system wants to converse upon, for example restaurant booking, flight booking etc.

\subsection{Training Joint Text and Knowledge Graph Embeddings}
Learning distributed embeddings for words has been widely studied for neural network based natural language understanding. \cite{mikolov2013distributed} was the first to propose a model to learn such distributed representations. In order to leverage additional information from knowledge graphs in goal-oriented dialogues, we need to first project the embeddings for both texts appearing in the dialogue corpus and knowledge graphs in the same vector space.
For learning such word embeddings, we adopted the work done by~\cite{Alsuhaibani2018}, which proposes a model for training the embeddings from both a corpus and a knowledge graph jointly. The proposed method is described in three steps: firstly training the Global Vectors (GloVe) following \cite{pennington2014}, secondly incorporating the knowledge graph, and jointly learning the former two using the proposed model with a linear combination of both their objective functions.
\subsubsection{Glove Vectors (GloVe)}
Firstly, we train the embedding vectors by creating a global word co-occurrence matrix $X$ from both the input and target sentences, where each word in these sentences is represented by a row in $X$ containing the co-occurrence of context words in a given contextual window. Finally, the GloVe embedding learning method minimizes the weighted least squares loss presented as:
\begin{equation}
J_C= \frac{1}{2}\sum_{i\in V}\sum_{j\in V}f(X_{ij})(w_i^\top w_j +bi +bj -log(X_{ij}))^2
\end{equation}
Where $X_{ij}$ denote the total occurrences of target word $w_i$ and the context word $w_j$ and the weighting function $f$ assigns a lower weight for extremely frequent co-occurrences to prevent their over-emphasis.
\subsubsection{Incorporating the knowledge graph}
The GloVe embeddings themselves do not take the semantic relation between corresponding words into account. 
Therefore, in order to leverage a knowledge graph while creating the word embeddings, we define an objective $J_s$ that only considers the three-way co-occurrences between a target word $w_i$ and one of its context word $w_j$ along with the semantic relation $R$ that exists between them in the KG. The KG-based objective is defined as follows:
\begin{equation}
J_S= \frac{1}{2}\sum_{i\in V}\sum_{j\in V}R(w_i,w_j)(w_i - w_j)^2
\end{equation}
Where $R(w_i, w_j)$ indicates the strength of the relation $R$ between $w_i$ and $w_j$, which assigns a lower weight for extremely frequent co-occurrences to prevent over-emphasising such co-occurrences and if there is no semantic relation between $w_i$ and $w_j$ then $R(w_i, w_j)$ will be set to zero.

For the final step, to train both the corpus and the KG objectives jointly. We define the combined objective function $J$ as their linearly weighted combination

\begin{equation}
J = J_C+\lambda J_S
\end{equation}

 $\lambda$ is the regularization coefficient that regulates the influence of the knowledge graph on those learned from the corpus.
 
 
\subsection{Regularizing using additional Entity Loss}

After projecting the knowledge graphs and the text in the same vector space, we compute an additional loss which is equal to the vector distances between the predicted entity and correct entity.
The intuition behind using this loss is to penalize the model for predicting the wrong entity during decoding, in vector space.
We use cosine distance to measure the entity loss as given by:
\begin{equation}
\text{Entity}_l = 1-\frac{\phi^{e_{corr}}.\phi^{e_{pred}}} {\lVert{\phi^{e_{corr}}}\rVert \lVert{\phi^{e_{pred}}}\rVert}
\end{equation}

\noindent
$\phi^{e_{corr}}$ and $\phi^{e_{pred}}$ being the vector embeddings for the correct and predicted entities, respectively.

\begin{figure}
    \centering
    \includegraphics[width=\textwidth]{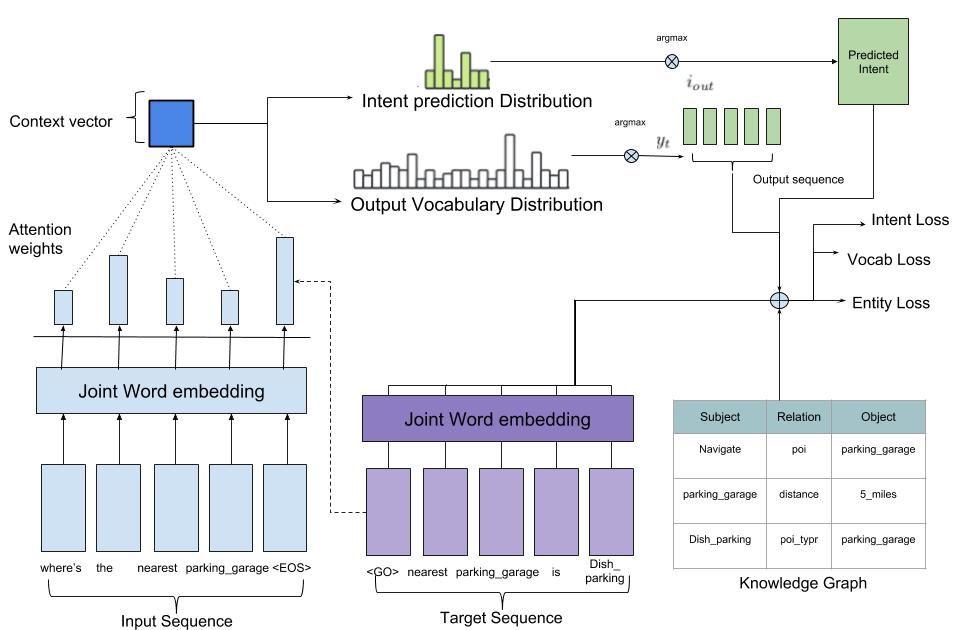}
        \vspace{0.1cm}
        
    \caption{Sequence-to-Sequence based module with multi-task learning of user intent regularized with additional Entity Loss}
    \label{fig:model}
    \vspace{-0.5cm}
\end{figure}

\subsection{Final objective Function}

The final objective function for training the model is the summation of the cross-entropy (CE) loss from the decoder ($\text{Vocab}_l$), the cross-entropy loss from the intent predictor ($\text{Intent}_l$) and the entity loss ($\text{Entity}_l$).

\noindent
The total loss is given by:
\begin{equation}
L_{tot} = \text{Vocab}_{l} + \text{Intent}_{l} + \text{Entity}_l
\end{equation}

\noindent
The model is trained with back-propagation using Adam \cite{kingma2014adam} optimizer.

\subsection{Key-Value Entity Look-up}
\label{kvl}
\begin{figure}
    \centering
    \includegraphics[width=\textwidth]{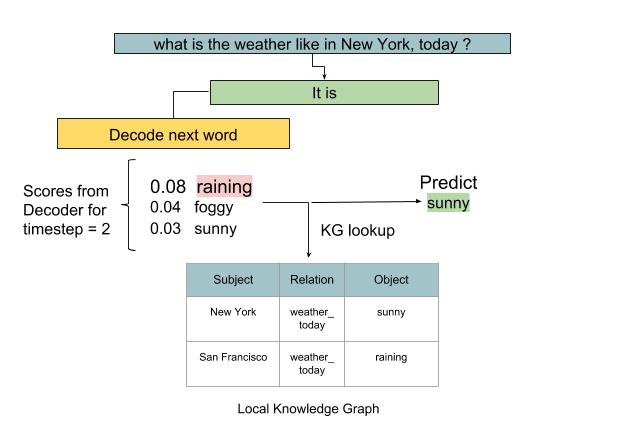}
    \vspace{-0.5cm}
    \caption{An example of how KVL works to replace predicted entities with correct ones from the local KG.}
    \label{fig:KVl}
\end{figure}

As mentioned before the knowledge graph (KG) can change from one dataset to another or even as in our case from one dialogue to another. This will result in the generation of KG entities that do not belong to the current dialogue. Since we are incorporating global knowledge information using the joint embedding, this additional step would ensure local KG integration aiding in the betterment of the models.
To accomplish this, we added an additional lookup step during decoding. While generating a new token during decoding at time-step $t$, we firstly check if the predicted token is an entity. If it's an entity, we first do a lookup into the local dialogue KG to check its presence. If the entity is not present, then we pick the one with the entity with the highest softmax probability that is present in the local KG using greedy search. The technique is illustrated in figure \ref{fig:KVl}. As seen in the figure, for the query \textit{what is the weather like in New York, today?}, during decoding at $t=2$, the model outputs a softmax distribution with highest probability\footnote{this is a fictitious example for explaining the algorithm, the scores are not what is being predicted from the real case scenarios} (0.08) for the token \textit{raining} after predicting \textit{it is}. We keep a copy of all the global KG entities and first check whether the predicted token is an entity in the global KG or not. Since \textit{raining} is indeed an entity, we further do a look up into the local dialogue KG if it exists. For this scenario it doesn't, hence we do a greedy search for the next best-predicted tokens which are present in the local KG. In this specific case, \textit{sunny} is the next most probable entity that exists in the local KG, We pick this entity for the time-step and feed it into the next. The final output will be \textit{it is sunny today in New York.} 

One such example in a real-case setting is shown in table \ref{tab:exampleNav}. The distance for the gas station was generated as \textit{2\_miles}, whereas we can observe that in the local KG, the correct distance entity is \textit{5\_miles}. Hence by performing the additional entity lookup, we replace \textit{2\_miles} with the correct entity during decoding. Empirical evidence suggests that this strategy can gain in performance as depicted in table \ref{tab:ablation}.

\section{Experiments}
\subsection{Dataset}
To test our hypothesis of multi-task learning of intent and dialogues, we needed a dataset which has multiple domains and user intents annotated along with a knowledge graph for task-oriented dialogues. \cite{eric-manning:2017:SIGDIAL} introduced such a multi-turn, multi-domain task-oriented dialogue dataset. They performed a Wizard-of-Oz based data collection scheme inspired by~\cite{wen2016network}. They used the Amazon Mechanical Turk (AMT) platform for data collection. The statics of the datasets are mentioned in Table~\ref{tab:stat}. The dataset is a conversation workflow in an in-car setting where the driver (user) is asking the assistant for scheduling appointments, weather forecasts, and navigation. Both roles are played by AMT turkers (workers). In driver mode, the turker asked a question while the assistant turker responds to the query using the provided knowledge graph. The Knowledge graph statistics are provided in Table \ref{tab:stat2}. 
\begin{table*}[ht]
    \caption{Statistics of the in-car multi-turn, multi-domain, goal-oriented dialogue dataset.}
    \centering  
     \begin{tabular}{ l | c}
     \toprule
     Training Dialogues & 2,425\\
     Validation Dialogues &	302\\
     Test Dialogues	& 304 \\
     Avg. \# Utterances Per Dialogue &	5.25 \\
     Avg. \# Tokens Per Utterance &	9 \\
     Vocabulary Size &	1,601 \\
     \bottomrule
    \end{tabular}
    \vspace{0.1cm}
    
    \label{tab:stat}
\end{table*}
\subsection{Pre-processing and Model Hyperparameters}
We processed the dataset by creating inputs out of the driver's utterances and the accompanied KG and the model outputs would be the current task intent and the assistants' output. We also processed the entire knowledge graph to identify the entities and convert them into canonicalized forms. We also split entities with multiple objects. For instance, a weather forecast that looks like "frost, low of 20F, high of 30F" can be split into (weather-condition="frost", low-temperature="low of 20F", high-temperature="high of 30F"). 

For training the Joint Embedding we used a contextual window for the co-occurrence matrix equal to 15 which is $\frac{1}{4}^{th}$ of the mean of input size. We trained the model with  $\alpha= 0.01$ and $\lambda= 10000$ and the embedding vectors with dimensions of 300 for 500 epochs with learning rate equals to $1e-4$ and a weight-decay of $1e-6$. The stats regarding the knowledge graph is provided in Table \ref{tab:stat2}.

As for the sequence-to-sequence model, we trained each of them on a GPU with 3072 CUDA cores and a VRAM of 12GB. The model is trained for 1000 epoch with a batch size of 128 and a hidden layer size equal to the embedding dimension of 300. We set the learning rate to be $1e-4$ for the encoder and $5e-4$ for the decoder, we also added a gradient clipping of 50.0 for countering the `exploding gradient` problem; by doing so we prevent the gradients from growing exponentially and either overflow (undefined values), or overshoot steep cliffs in the cost function. We didn't do any exhaustive hyper-parameter optimization, the values reported here are the ones used for all models. Our codes and preprocessing scripts are available in \footnote{https://github.com/s6fikass/Chatbot\_KVNN}.


\begin{table*}[ht]
    \centering  
    \caption{Statistics of the trained Joint Embedding.}
     \begin{tabular}{ l | c}
     \toprule
     Number of entities & 291 \\
     Number of Triples & 2,512\\
     Number of Triples found in context & 1,242\\
     Max Triple co-occurrence  &	750\\
     Unique words co-occurrences	& 1,30,803 \\
     \bottomrule
    
    \end{tabular}
       \vspace{0.1cm}

    \label{tab:stat2}
\end{table*}

\subsection{Results}
In this section, we report the results from our model on the mentioned dataset. We evaluate our system using BLEU scores as suggested by \cite{papineni2002bleu}. BLEU as defined by \cite{papineni2002bleu} analyzes the co-occurrences of n-grams in the reference and the proposed responses. It computes the n-gram precision for the whole dataset, which is then multiplied by a brevity penalty to penalize short translations We are reporting the geometric means of BLEU-1, BLEU-2, BLEU-3, and BLEU-4. In addition to that we evaluated the both our model and the Mem2Seq model proposed by \cite{P18-1136} using an Embedding-based metrics such as Greedy Matching, Embedding Average and Vector Extrema as proposed by \cite{liu2016not} \footnote{We report these metrics for our best model and only for Mem2Seq because their implementation is open-source.}. 
We are using an Attention based seq-to-seq model as a baseline along with the other state-of-the-art models on this corpora \cite{eric-manning:2017:SIGDIAL} (KV Retrieval Net) and \cite{P18-1136} (Mem2Seq H3). We are reporting our best model which is attention based Sequence-to-sequence model (S2S+Intent+JE+EL) with Joint Embeddings (JE) as inputs and trained as a multi-task objective of learning response utterance along with user Intent further, regularized using Entity Loss (EL). This model is further improved with an additional step called Key-Value Entity Look-up (KVL) which is done during inference. The method was explained in \ref{kvl}.

\begin{table*}[ht]
\centering
\caption{Results compared to baseline and State-of-the-art models.}
    \begin{tabular}{l|l|l|l|l}
    \toprule
        \textbf{Model} &  \textbf{BLEU}  &  \textbf{Emb. Avg.} &  \textbf{Vec. Ext.} &  \textbf{Greedy}\\
    \hline
        Attention based Seq-to-Seq  & 8.4 & - & - & -  \\
    \hline
        KV Retrieval Net(no enc attention)  &  10.8 & - & - & - \\
        KV Retrieval Net &  13.2 & - & - & - \\
        Mem2Seq H3 & 12.6 & 0.668 & 0.742 & 0.528 \\
    \hline
        S2S+Intent+JE+EL &  \textbf{14.12} & - & - & - \\
        S2S+Intent+JE+EL+KVL &  \textbf{18.31} & \textbf{0.955} & \textbf{0.974} & \textbf{0.625} \\
    \bottomrule
    \end{tabular}
     \vspace{0.1cm}
    \label{tab:eval}
\end{table*}

As seen from the results, our proposed architecture has absolute improvements of 0.92 over KV Retrieval Net and 1.52 over Mem2Seq (H3) on BLEU. Although, the results are not directly comparable with the latter because we use canonicalized entity forms like \cite{eric-manning:2017:SIGDIAL}. We are reporting BLEU scores because for task-oriented dialogues there's not much variance between the generated answers, unlike open-domain dialogues \cite{liu2016not}. We are using the official version (i.e. moses  multi-bleu.perl script)
\footnote{\raggedright multi-bleu: \url{https://raw.githubusercontent.com/moses-smt/mosesdecoder/master/scripts/generic/multi-bleu.perl} }for doing all the evaluations. Also, as seen in table \ref{tab:eval}, our proposed model performs significantly better than Mem2Seq on emedding-based metrics too.

\begin{table*}[ht]
    \centering
    \caption{Ablation Study.}
    \begin{tabular}{l|l|l}
        \toprule
        \textbf{Model Used} &  \textbf{BLEU Score} & \textbf{BLEU with KVL} \\
        \hline
        S2S+glove &  10.42 & 14.63 \\
        S2S+JE &  13.35 & 15.29 \\
        S2S+Intent+JE & 12.65 & 17.89  \\
        S2S+Intent+glove &  13.25 & 17.27 \\
        S2S+Intent+JE+EL &  \textbf{14.12} & \textbf{18.31} \\

        \bottomrule
    \end{tabular}
     \vspace{0.1cm}
    \label{tab:ablation}
\end{table*}


\section{Ablation Study}

As mentioned in the results, our best model is based on Joint embeddings as inputs, followed by jointly learning user intent and text generation; with entity loss based on further regularization techniques. To analyze which specific part is influencing the performance, we did an ablation study dissecting the different components of the model one by one.
The performances are reported in Table \ref{tab:ablation}.

The first model is a sequence-to-sequence (S2S) model with Glove \cite{pennington2014} embeddings as input. These embeddings are trained on the training set. It can be seen that with the domain-vocabulary knowledge the model already performs better than vanilla attention sequence-to-sequence model. The next (S2S+JE) is the same model with Joint text and knowledge graph embeddings (also trained on the dialogue corpus and provided knowledge graph). This model has an absolute improvement of 2.93 in BLEU scores. Interestingly, adding the intent along with Joint Embeddings drops the performance by 0.7 but the encounters a little improvement with glove vectors. The model sees a further boost in performances (relative improvement of 3.7) over vanilla models with GloVe. All the models encounter better performances with the proposed key-value lookup (KVL) technique during inferencing.

\section{Qualitative Analysis}
 
 To qualitatively understand the performance improvements of the model with a different setting as in Table \ref{tab:ablation} we analyze the outputs from all of them for a given user query. For the user (driver) query \textit{Where is the nearest gas station ?}, predicted responses are given in Table \ref{tab:exampleNav}. The knowledge graph snapshot for this particular dialogue scenario is in Table \ref{tab:kb_nav}.
 
\begin{table*}
\centering
\caption{Example of generated responses on the Navigation domain explaining the improvement process through the model stages.}
\begin{tabular}{l|l}

\hline
    \textbf{Driver} & Where is the nearest gas station? \\
\hline
    S2S & chevron is 5\_pm away at <UNK> it is going to snow \\
    S2S+GloVe &  chevron is 3\_pm at room\_215 \\
    S2S+JE & chevron is 5\_miles at <UNK> high\_of\_30\_f \\
    S2S+Intent+JE & chevron is at 783\_arcadia\_pl\\
    S2S+Intent+JE+EL & chevron is 2\_miles away at 783\_arcadia\_pl \\ 
    S2S+Intent+JE+EL+KVL & chevron is 5\_miles away at 783\_arcadia\_pl \\
\hline
    Target & chevron is 5\_miles away at 783\_arcadia\_pl \\
\hline

\end{tabular}
 \vspace{0.1cm}

 \label{tab:exampleNav}
\end{table*}

 As observed, the sequence-to-sequence (S2S) model is not able to produce knowledge grounded responses since it has no information about the background knowledge graph. Using Joint Embeddings (S2S+JE), although produces more knowledge grounded responses (since chevron is an entity of type gas\_station and is at a distance of 5\_miles), it outputs entities like high\_of\_30\_f which is not related to navigation intent but the weather. Incorporating intent into the model (S2S+Intent+JE) ensures that it is generating words conditionally dependent on the intent also. Further grammatical improvements in the model response are seen with entity loss (S2S+Intent+JE+EL) which is further made more knowledge grounded with the proposed KVL method as seen in the last response (S2S+Intent+JE+EL+KVL).
 To further understand the quality of the produced responses, we did a human evaluation of a subset of the predicted responses from the test set. We asked the annotators to judge the response whether it is human-like or not on a scale of 1-5. The average score given by the annotators is 4.51.
 
\begin{table}[ht]
\centering
\small
\caption{KG triples for the dialogue in Table \ref{tab:exampleNav}, for navigation.}
\begin{tabular}{l|l|l}
\hline
    \textbf{Subject}  & \textbf{Relation}  & \textbf{Object}
  \\ \hline
    chevron & distance & 5\_miles \\ 
    chevron & traffic\_info & moderate\_traffic  \\ 
    chevron & poi\_type & gas\_station \\chevron & address & 783\_arcadia\_pl \\

    \hline

\end{tabular}
 \vspace{0.1cm}
\label{tab:kb_nav}
\end{table}

\section{Error Analysis}

To further understand the model performance, we did a further analysis of the responses from the test set which gave very low BLEU scores during evaluation. The target and the predicted response are in Table \ref{tab:ErrorA}.

\begin{table}[ht!]
\small
\centering
\caption{Error Analysis.}
    \begin{tabular}{p{6cm}|p{6cm}}
        \toprule
        \textbf{Predicted Sentence} &  \textbf{Target Sentence} \\
        \hline
        
        \text{You're welcome } & \text{Have a good day } \\
        \text{What city do you want the forecast for} & \text{What city can i give you this weather for } \\
        \text{Setting gps for quickest route now } & \text{I picked the route for you drive carefully}  \\
        \text{It is cloudy with a low of 80f in fresno} & \text{There are no clouds in fresno right now} \\

  \bottomrule

    \end{tabular}
     \vspace{0.1cm}
    \label{tab:ErrorA}    
\end{table}

As observed, the model produces fewer word overlaps for generic responses like \textit{have a good day}. But, the responses are grammatically and semantically correct, which cannot be measured using BLEU scores. The other types of errors are factual errors where the model fails because it requires reasoning as in case of the $4^{th}$ example. The user is asking if the weather is cloudy in this case, in Fresno. 
\begin{table}[ht]
\centering
\small
\caption{KG triples and context for the error analysis in Table \ref{tab:ErrorA} for weather.}
\begin{tabular}{l|l|l}
\hline
    \textbf{Subject}  & \textbf{Relation}  & \textbf{Object}
  \\ \hline
        fresno & monday & clear\_skies \\
        fresno & monday & low\_40f \\


\hline
\end{tabular}
 \vspace{0.1cm}

\label{tab:kb_f_nav}
\end{table}

\section{Conclusions and Future Work}

In this paper, we proposed techniques to improve goal-oriented dialogue systems using joint text and knowledge graph embeddings with learning user query intent as a multi-task learning paradigm. We also further suggested regularization techniques based on entity loss to improve upon the model. Empirical evaluations suggest that our suggested method can gain over existing state-of-the-art systems on BLEU scores for multi-domain, task-oriented dialogue systems. We are not reporting entity-f1 scores since we are treating canonicalized entity forms as vocabulary tokens and BLEU scores will already reflect the presence of the correct entity in the responses.

For future endeavors, we would like to include KVL in the training using memory network modules instead of using it as a separate module during inference. We also observed that our model can benefit from better out-of-vocabulary (OOV) words handling which we would like to keep as a future work. Also, as observed in predicted responses for generic dialogues in first 3 rows in table \ref{tab:ErrorA}, it would make sense to incorporate evaluations which also captures semantic similarities between predicted responses, we would like to work on this in future. 

\section{Acknowledgements}

This work was partly supported by the European Union`s Horizon 2020 funded projects WDAqua (grant no.~642795) and Cleopatra (grant no.~812997) as well as the BmBF funded project Simple-ML.

%
%
%
%





\bibliographystyle{splncs04}
\bibliography{bibliography}
\end{document}